\documentclass[11pt,a4paper]{article}
\usepackage[hyperref]{acl2019}
\usepackage{times}
\usepackage{latexsym}
\usepackage{amsmath,amssymb}
\usepackage{graphicx}
\usepackage{url}
\usepackage{xspace}
\usepackage{wrapfig,lipsum,booktabs}

\aclfinalcopy 
\def\aclpaperid{***} 

\usepackage{array}
\newcolumntype{H}{>{\setbox0=\hbox\bgroup}c<{\egroup}@{}}

\newcommand{\system}{{{\textsc{DialoGPT}}}\xspace}

\newcommand{\neurocon}{{{\textsc{PersonalityChat}}}\xspace}

\title{\textsc{DialoGPT} : Large-Scale Generative Pre-training\\ 
for Conversational Response Generation}
\author{Yizhe Zhang \quad\quad Siqi Sun \quad\quad Michel Galley \quad\quad Yen-Chun Chen \\ \textbf{Chris Brockett} \quad\quad\textbf{Xiang Gao}\quad\quad \textbf{Jianfeng Gao}\quad\quad \textbf{Jingjing  Liu} \quad\quad \textbf{Bill Dolan}\\
    Microsoft Corporation, Redmond, WA, USA \thanks{\quad 
    A collaboration between Microsoft Research and Microsoft Dynamics 365 AI Research. } \\
  {\small \tt \{yizzhang,siqi.sun,mgalley,yenchen,chrisbkt,xiag,jfgao,jingjl,billdol\}@microsoft.com}
}

\begin{document}
\maketitle

\begin{abstract}
We present a large, tunable neural conversational response generation model, \system (dialogue generative pre-trained transformer).
Trained on 147M conversation-like exchanges extracted from Reddit comment chains over a period spanning from 2005 through 2017, 
DialoGPT extends the Hugging Face PyTorch transformer to attain a performance close to human both in terms of automatic and human evaluation in single-turn dialogue settings. 
We show that conversational systems that leverage DialoGPT generate more relevant, contentful and context-consistent responses than strong baseline systems.  
The pre-trained model and training pipeline are publicly released to facilitate research into neural response generation and the development of more intelligent open-domain dialogue systems.

\end{abstract}

\section{Introduction}

We introduce  \system, a tunable gigaword-scale neural network model for generation of conversational reponses, trained on Reddit data. 

Recent advances in large-scale pre-training using transformer-based architectures \cite{gpt2, devlin2019bert, 2019t5} have achieved great empirical success. 
OpenAI's GPT-2 \cite{gpt2}, for example, has demonstrated that transformer models trained on very large datasets can capture long-term dependencies in textual data and generate text that is fluent, lexically diverse, and rich in content. 
Such models have the capacity to capture textual data with fine granularity and produce output with a high-resolution that closely emulates real-world text written by humans. 

\system extends GPT-2 to address the challenges of conversational neural response generation. Neural response generation is a subcategory of text-generation that shares the objective of generating natural-looking text (distinct from any training instance) that is \textit{relevant} to the prompt. 
Modelling conversations, however, presents distinct challenges in that  human dialogue, which encapsulates the possibly competing goals of two participants, is intrinsically more diverse in the range of potential responses \cite{li2015diversity, zhang2018generating,gao2019neural, jointly2019gao}. It thus poses a greater \textit{one-to-many} problem than is typical in other text generation tasks 
such as neural machine translation, text summarization and paraphrasing.
Human conversations are also generally more informal, noisy, and, when in the form of textual chat,  often contain informal abbreviations or syntactic/lexical errors.  

Most open-domain neural response generation systems suffer from content or style inconsistency \cite{li2016persona,zhang2019consistent,gao2019structuring}, 
lack of long-term contextual information \cite{serban2017hierarchical}, and 
blandness \cite{li2015diversity,zhang2018generating,qin2019conversing}. While these issues can be alleviated by modelling strategies specifically designed to boost information content, a transformer-based architecture like GPT-2 \cite{gpt2}, which uses a multi-layer self-attentive mechanism to allow fully-connected cross-attention to the full context in a computationally efficient manner, seems like a natural choice for exploring a more general solution.
Transformer models, for example, allow long-term dependency information to be better be preserved across time \cite{gpt2}, thereby improving content consistency. 
They also have higher model capacity due to their deep structure (up to 48 layers in GPT-2) and are more effective in leveraging large-scale datasets (more than 100 million training instances) than RNN-based approaches \cite{vaswani2017attention}. 

Like GPT-2, \system is formulated as an autoregressive (AR) language model, and uses the multi-layer transformer as model architecture. 
Unlike GPT-2, however, \system is trained on large-scale dialogue pairs/sessions extracted from Reddit discussion chains. 
Our assumption is that this should enable \system to capture the joint distribution of $P(\text{Target},\text{Source})$ in conversational flow with finer granularity. 
In practice, this is what we observe: sentences generated by \system are diverse and contain information specific to the source prompt, analogous what GPT-2 generates for continuous text. 
We have evaluated the pre-trained model on a public benchmark dataset (DSTC-7), and a new 6k multi-reference test dataset extracted from Reddit postings. 
\system achieves state-of-the-art results in both automatic and human evaluation, lifting performance to near-human response quality.

We have released the source code and a pre-trained model to facilitate future research.\footnote{GitHub: \url{https://github.com/microsoft/DialoGPT}; Blog: \quad \url{https://aka.ms/dialogpt}}. Our model can be easily leveraged and adapted to new dialogue datasets, especially datasets with few training examples.  The \system package also contains an open-source training pipeline (data extraction/preparation and model training/evaluation) built upon the Huggingface PyTorch transformer \cite{pytorchtransformer}. \footnote{Our model is also available over Hugging face Transformers. \url{https://huggingface.co/microsoft/DialoGPT-medium}}

%

\section{Dataset}

The dataset is extracted from comment chains scraped from Reddit spanning from 2005 till 2017.
Reddit discussions can be naturally expanded as tree-structured reply chains, since a thread replying to one thread forms the root node of subsequent threads.  We extract each path from the root node to the leaf node as a training instance containing multiple turns of dialogue.

We filter the data by removing the instances where (1) there is a URL in source or target, (2) where the target contains word repetitions of at least three words, (3) where the response does not contain at least one of the top-50 most frequent English words (e.g., ``the'', ``of'', ``a''), since this probably indicates it might not be an English sentence, 
(4) where the response contains special markers such as ``['' or ``]'', as this could be markup language,
(5) where source and target sequences together are longer than 200 words,
(6) where the target contains offensive language, identified by phrase matching against a large blocklist.
We also excluded a large number of subreddits that had been identified as likely to contain offensive content.  
In addition, we aggressively filtered out blandness, e.g., removing instances where the responses contained 90\% of tri-grams that have been seen more than 1000 times. Often uninformative, such responses account for about 1\% of the data.
After filtering, the dataset comprises 147,116,725 dialogue instances, in total 1.8 billion words.

\section{Method}

\subsection{Model Architecture}

We trained our \system model on the basis of the GPT-2~\cite{gpt2} architecture.
The GPT-2 transformer model adopts the generic transformer language model~\cite{vaswani2017attention} and leverages a stack of masked multi-head self-attention layers to train on massive web-text data. The text generated either from scratch or based on a user-specific prompt is realistic-looking. 
The success of GPT-2 demonstrates that a transformer language model is able to characterize human language data distributions at a fine-grained level, presumably due to large large model capacity and superior efficiency.

Our model inherits from GPT-2 \cite{gpt2}, a 12-to-48 layer transformer with layer normalization, a initialization scheme that accounts for model depth that we modified, and byte pair encodings \cite{bpe} for the tokenizer. 
We follow the OpenAI GPT-2 to model a multi-turn dialogue session as a long text and frame the generation task as language modeling. 
We first concatenate all dialog turns within a dialogue session into a long text ${x_1, \cdots , x_N}$ ($N$ is the sequence length), ended by the end-of-text token. 
We denote the source sentence (dialogue history) as $S={x_1, \cdots , x_m}$ and target sentence (ground truth response) as $T={x_{m+1}, \cdots , x_N}$, the conditional probability of $P(T|S)$ can be written as the product of a series of conditional probabilities:
\begin{align}\label{eq:lm}
	p(T|S) = \prod_{n=m+1}^N p(x_n|x_1, \cdots, x_{n-1})
\end{align}
For a multi-turn dialogue session ${T_1, \cdots, T_K}$, \eqref{eq:lm} can be written as $p(T_K,\cdots,T_2|T_1)$, which is essentially the product of conditional probabilities of $p(T_i|T_1,\cdots,T_{i-1})$. 
Consequently, optimizing a single objective $p(T_K,\cdots,T_2|T_1)$ can be perceived as optimizing all $p(T_i|T_1,\cdots,T_{i-1})$ source-target pairs.

 Our implementation is based on the open-source PyTorch-transformer repository.\footnote{\url{https://github.com/huggingface/pytorch-transformers}}

\subsection{Mutual Information Maximization}
\label{sec:mmi}
Open-domain text generation models are notorious for generating bland, uninformative samples. To address this problem, we implement a maximum mutual information (MMI) scoring function~\cite{li2015diversity, zhang2018generating}. MMI employs a pre-trained \textit{backward} model to predict source sentences from given responses, i.e., $P(\text{Source}|\text{target})$. We first generate a set of hypotheses using top-K sampling. Then we use the probability of $P(\text{Source}|\text{Hypothesis})$ to rerank all hypotheses. Intuitively, maximizing backward model likelihood penalizes the bland hypotheses, as frequent and repetitive hypotheses can be associated with many possible queries, thus yielding a lower probability for any specific query.

We also attempted to optimize the reward $R \triangleq P(\text{Source}|\text{Hypothesis})$ using a policy gradient \cite{williams1992simple} with a sample-averaged baseline, following \citet{zhang2018generating}. The validation reward can be stably improved, but unlike the training under RNN architecture, we observed that reinforcement learning (RL) training  easily converges to a degenerate locally-optimal solution, where the hypothesis simply repeats the source sentence (i.e., a parroting model) and mutual information is maximized. We hypothesize that transformers can become trapped in local optima due to their strong model representation power. 
We leave the investigation of regularized RL training to future work.

\section{Result}

\subsection{Experimental Details}
We trained 3 different sizes of the model with total parameters of 117M, 345M and 762M  respectively. 
The model specification follows \citet{gpt2} (Table~\ref{tab:param}).

\begin{table}[ht!]
\centering
\begin{tabular}{l|c|c|c}
Model &Layers & $D_{emb}$ & B\\
\hline
117M& 12 & 768 & 128\\
345M& 24 & 1024 & 64\\
762M& 36 & 1280 & 32\\
\end{tabular}
\vspace{1mm}
\caption{Model configurations. ``B'' denotes batch size per GPU. }
\label{tab:param}
\end{table}

Our model uses a vocabulary of 50,257 entries, and was trained on 16 Nvidia V100 machines with NVLink. 
We used the Noam learning rate scheduler with 16000 warm-up steps. The learning rate is selected based on validation loss. Each model is trained until there is no progress in validation loss. For small and medium models, we trained the models for up to 5 epochs. For the large model we trained for at most 3 epochs. 

\paragraph{Speeding up training}
To accelerate the training process and accommodate GPU memory limitations, we first compress all training data into a lazy-loading database file, so that data is loaded only when needed (pre-fetching large chunks to reduce access frequency). We also leverage separate asynchronous data processes to scale the training. As a result, training time declines  approximately linearly w.r.t. the number of GPUs. 
We further employed a dynamic batching strategy to group conversations of similar lengths into the same batch, thus increasing training throughput.  

\subsection{DSTC-7 Dialogue Generation Challenge}

The DSTC (Dialog System Technology Challenges) 7 track \cite{galley2019grounded} is an end-to-end conversational modeling task,\footnote{\url{https://github.com/mgalley/DSTC7-End-to-End-Conversation-Modeling/tree/master/evaluation}} in which the goal is to generate conversation responses that go beyond chitchat by injecting information that is grounded in external knowledge. 
This task is distinct from what is commonly thought of as goal-oriented, task-oriented, or task-completion dialogs in that there is no specific or predefined goal (e.g., booking a flight, or reserving a table at a restaurant). 
Instead, it targets human-like interactions where the underlying goal is often ill-defined or unknown in advance, of the kind seen in work and other productive environments (e.g., brainstorming meetings) where people share information.

The DSTC-7 test data contains conversation threads from Reddit data. In order to create a multi-reference test set, we utilized conversation sessions that contain 6 or more responses. 
Given other filtering criteria such as turn length, this yields a 5-reference test set of size 2208. (For each instance, one of the 6 human responses is set aside to assess human performance on this task.) Note that our training data is collected from a different time span from the test set.
We performed automatic evaluation using standard machine translation metrics, including BLEU \cite{papineni2002bleu}, METEOR \cite{lavie2007meteor}, and NIST \cite{doddington2002nist}. 
NIST is a variant of BLEU that weights n-gram matches by their information gain, i.e., it indirectly penalizes uninformative n-grams. 
We also use Entropy \cite{zhang2018generating} and Dist-n \cite{li2015diversity} to evaluate lexical diversity.
More details are provided in \citet{galley2019grounded}. 

\begin{table*}[ht!]
\small
\centering
\begin{tabular}{r H  r H  r | H r  H r | r| r  |r  r | r}
\cmidrule[\heavyrulewidth]{1-14}

 & \multicolumn{4}{c|}{NIST} & \multicolumn{4}{c|}{BLEU} & METEOR & Entropy & \multicolumn{2}{c|}{Dist} & \multicolumn{1}{c} {Avg Len}  \\ 
Method & N-1 & N-2 & N-3 & N-4 & B-1 & B-2 & B-3 & B-4 &  & E-4 &  D-1 &D-2 & \\
\cmidrule[\heavyrulewidth]{1-14} 
\neurocon & 0.18 & 0.19 & 0.20 & 0.20 & 31.73\% & 10.44\% & 3.92\% & 1.47\% & 5.42\% & 6.89 & 5.9\% & 16.4\% & 8.2 \\
Team B & 2.34 & 2.51 & 2.52 & 2.52 & 41.22\% & 14.35\% & 5.01\% & 1.83\% & 8.07\% & 9.03 & 10.9\% & 32.5\% & 15.1\\
\cmidrule[\heavyrulewidth]{1-14}
\system (117M)  & 1.49 & 1.58 & 1.59 & 1.60 & 34.09\% & 10.36\% & 4.28\% & 2.02\% & 7.17\% & 6.94 & 6.2\% & 18.94\% & 13.0\\
GPT(345M) & 1.70 & 1.78 & 1.79 & 1.79 & 33.41\% & 9.13\% & 2.86\% & 1.06\% & 6.38\% & 9.72 & 11.9\% & 44.2\% & 14.7\\
\system (345M) & 2.59 & 2.80 & 2.82 & 2.82 & 39.27\% & 14.16\% & 5.55\% & 2.31\% & 8.51\% & \textbf{10.08} & 9.1\% & 39.7\% & 16.9\\
\system (345M,Beam) & \textbf{2.59} & \textbf{2.92} & \textbf{2.96} & \textbf{2.97} & \textbf{42.38}\% & \textbf{19.18}\% & \textbf{10.27}\% & \textbf{6.05}\% & \textbf{9.29}\% & 9.57 & \textbf{15.7}\% & \textbf{51.0}\% & 14.2\\
\cmidrule[\heavyrulewidth]{1-14}
Human & 2.42 & 2.62 & 2.65 & 2.65 & 34.08\% & 12.35\% & 5.72\% & 3.13\% & 8.31\% & 10.45 & 16.7\% & 67.0\% & 18.8\\
	\cmidrule[\heavyrulewidth]{1-14}
	\end{tabular}
\caption{DSTC evaluation. ``Team B'' is the winner system of the DSTC-7 challenge. ``Beam'' denotes beam search.  ``Human'' represents the held-out ground truth reference.}\label{tab:dstc}
\end{table*}

\begin{table*}[ht!]
\small
\centering
\begin{tabular}{r H  r H  r | H r  H r | r| r  |r  r | r}
\cmidrule[\heavyrulewidth]{1-14}

 & \multicolumn{4}{c|}{NIST} & \multicolumn{4}{c|}{BLEU} & METEOR & Entropy & \multicolumn{2}{c|}{Dist} & \multicolumn{1}{c} {Avg Len}  \\ 
Method & N-1 & N-2 & N-3 & N-4 & B-1 & B-2 & B-3 & B-4 &  & E-4 &  D-1 &D-2 &  \\
\cmidrule[\heavyrulewidth]{1-14} 
\neurocon & 0.73 & 0.78 & 0.79 & 0.79 & 32.78\% & 11.22\% & 4.35\% & 1.95\% & 6.93\% & 8.37 & 5.8\% & 18.8\% & 8.12\\
\cmidrule[\heavyrulewidth]{1-14}
\multicolumn{13}{l}{\textit{Training from scratch:}}\\
\system (117M) & 1.16 & 1.23 & 1.30 & 1.37 & 34.69\% & 9.74\% & 3.80\% & 1.77\% & 6.17\% & 7.11 & 5.3\% & 15.9\% & 9.41\\
\system (345M) & 2.23 & 2.51 & 2.79 & 3.08 & 35.23\% & 16.92\% & 8.34\% & 4.59\% & 9.34\% & 9.03 & 6.7\% & 25.6\% & 11.16\\
\system (762M) & 2.24 & 2.52 & 2.81 & 3.10 & 42.53\% & 17.87\% & 9.07\% & 5.19\% & 9.53\% & 9.32 & 7.5\% & 29.3\% & 10.72\\
\cmidrule[\heavyrulewidth]{1-14}
\multicolumn{13}{l}{\textit{Training from OpenAI GPT-2:}}\\
\system (117M) & 2.25 & 2.39 & 2.41 & 2.41 & 35.43\% & 10.54\% & 3.85\% & 1.55\% & 7.53\% & 10.77 & 8.6\% & 39.9\% & 12.82\\
\system (345M) & 2.67 & 3.00 & 3.05 & 3.06 & 40.97\% & 16.96\% & 8.31\% & 4.56\% & 9.81\% & 9.12 & 6.8\% & 26.3\% & 12.19\\
\system (345M, Beam)  & 2.94 & \textbf{3.4} & \textbf{3.49} & \textbf{3.5} & \textbf{45.27}\% & \textbf{21.76}\% & \textbf{12.51}\% & \textbf{7.92}\% & 10.74\% & 10.48 & \textbf{12.38}\% & \textbf{48.74}\% & 11.34 \\
\system (762M) & 2.51 & 2.84 & 2.89 & 2.90 & 44.45\% & 18.66\% & 9.32\% & 5.25\% & 9.66\% & 9.72 & 7.76\% & 29.93\% & 11.19\\
\system (762M, Beam) & 2.51 & 2.90 & 2.97 & 2.98 & 44.84\% & 21.08\% & 12.01\% & 7.57\% & 10.11\% &  10.06  & 11.62\% & 44.07\% & 10.68\\
\system (345M, MMI) & \textbf{2.95} & 3.28 & 3.32 & 3.33 & 39.18\% & 15.68\% & 7.45\% & 3.94\% & \textbf{11.23}\% &  \textbf{11.25}  & 9.39\% & 45.55\% & 17.21\\
\cmidrule[\heavyrulewidth]{1-14}
Human & 2.99 & 3.41 & 3.83 & 4.25 & 39.61\% & 17.90\% & 10.71\% & 7.48\% & 10.64\% & 10.99 & 14.5\% & 63.0\% & 13.10\\
	\cmidrule[\heavyrulewidth]{1-14}
	\end{tabular}
\caption{6K Reddit multi-reference evaluation. ``Beam'' denotes beam search.  ``Human'' represents the held-out ground truth reference.}\label{tab:multiref}
\end{table*}

We compared \system with our in-house competitive sequence-to-sequence model \neurocon based on \cite{li2015diversity} and trained on Twitter data, which has been used in production as a Cognitive Service for Microsoft Azure.\footnote{Project \neurocon : \url{https://docs.microsoft.com/en-us/azure/cognitive-services/project-personality-chat/overview}}
Table~\ref{tab:dstc} summarizes the automatic evaluation results. \system with 345M parameters and beam search achieved the highest automatic score across most metrics.  Scores for \system with 345M parameters are better across the board than with 117M parameters. Beam search (with beam width 10) dramatically improves BLEU and DIST scores, and marginally improves NIST and METEOR. Note that our model is fine-tuned on source-target pairs, and does not leverage grounding information from the DSTC training set. Presumably, the model learns background information during pre-training and is unhindered by the lack of a grounding document. 
\begin{figure}[ht!]
    \centering
    \includegraphics[width=0.48\textwidth]{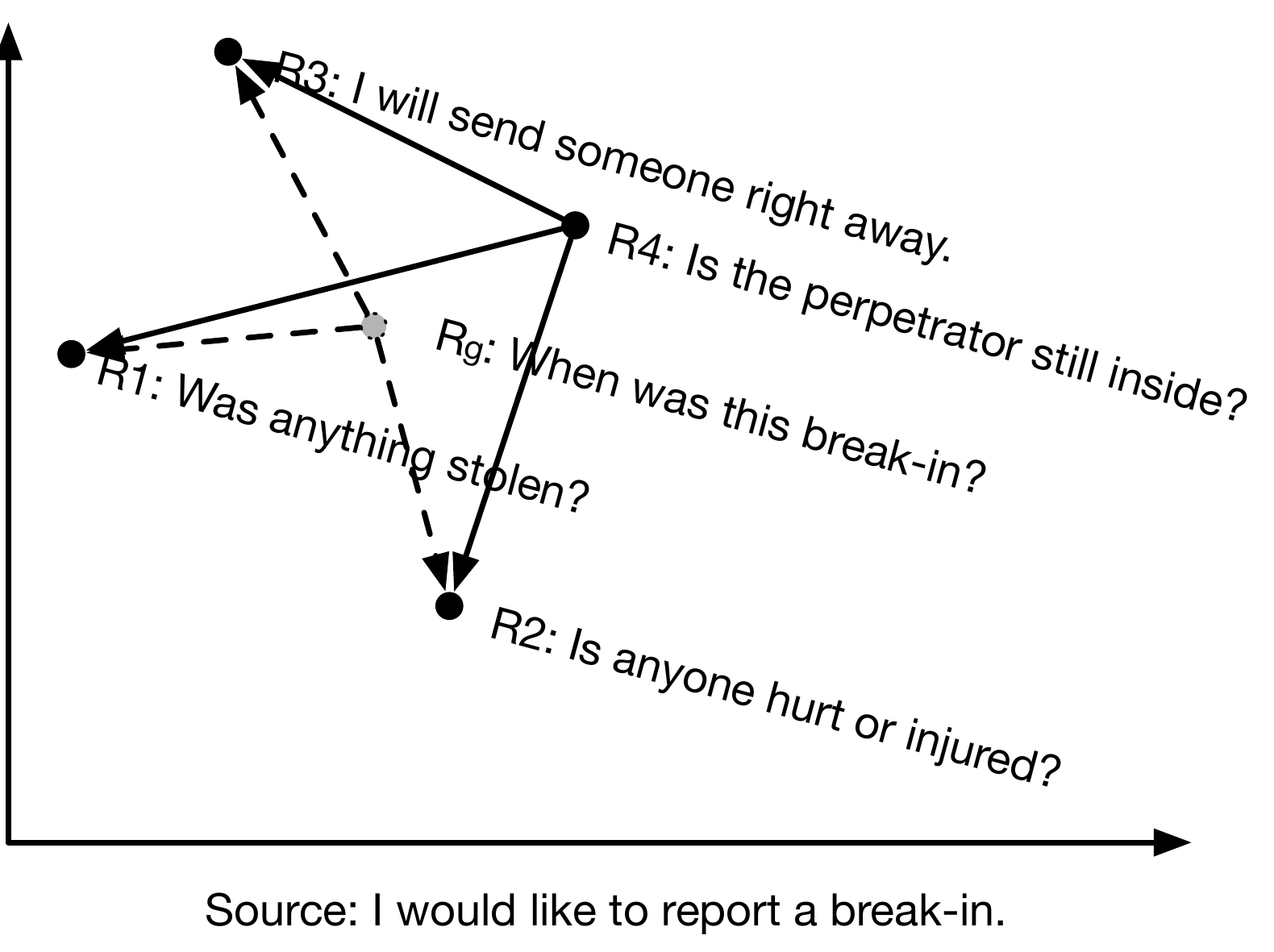}
    \caption{A generated response can surpass a human response in automatic metrics. Example responses are from \citet{gupta2019investigating}}
    \label{fig:emb_space}
\end{figure}

The automatic scores of \system are higher than those for humans. This does not mean that the generation is more ``realistic'' than human, but is probably attributable  to the one-to-many nature of conversation. As illustrated in Figure~\ref{fig:emb_space}, multiple human responses (R1-R4) can correspond well to a source utterance. Without loss of generality, suppose R1-R3 are the ``ground truth'' references that will be tested on, while R4 is the ``held-out'' human response that serves to compute a ``human'' score. 
In semantic space, a generated response $R_{g}$ from a well-trained model will presumably tend to lie in the vicinity the geometric center of all possible responses, because the training objective seeks to generate the most likely response. This may be close to the geometric mean of all training instances, thus ``averaging out'' these instances. Consequently, a generated response $R_g$ might have a lower ``semantic distance'' (manifested in higher automatic scores like BLEU) from R1-R3 than the targeted human response R4.

\subsection{A New Reddit Multi-reference Dataset}

We further evaluate \system on a multi-reference test set with 6K  examples. The results are shown in Table~\ref{tab:multiref}.
We test our method on two settings: training from scratch and fine-tuning using GPT-2 as the pre-trained model. In both settings, a larger model consistently outperforms a smaller one. Comparing training from scratch to fine-tuning from the pre-trained GPT-2 model, when applying to smaller model, using GPT-2 model gives larger performance gains.
Again, the best system \system  (345M, w/ beam search) scores higher on BLEU than humans. Larger models trained from scratch (345M and 762M) perform comparably to one finetuned on GPT-2.

\subsection{Re-ranking The Response Using MMI}
We perform mutual information maximization as described in Section~\ref{sec:mmi}. Specifically, we generate 16 samples for each input source sentence by using top-K sampling ($K=10$) using the 345M model fine-tuned from the GPT-2 medium model. This is followed by a re-ranking step using a backward model, which is also a 345M model fine-tuned from the GPT-2 medium model. The response that yields lowest backward model loss is selected for evaluation. The results are summarized in the second to last row in Table~\ref{tab:multiref}. 
It can be seen that compared with Greedy generation, MMI re-ranking produces more diverse responses with higher NIST, METEOR and Entropy and Dist scores, but with a slight drop in BLEU.

\subsection{Generation Examples}
We provide sample generated dialogues in Table~\ref{tab:generation} (interactive chat) and Table~\ref{tab:generation_2} (a self-playing bot with user prompt). The output is based on top-K sampling.\footnote{Owing to occasional toxic outputs, we are unable to release the generation code at this time.} Interestingly, our model exhibits the ability to address commonsense questions to some extent, presumably 
owing to the rich amount of information that can be learned from Reddit data. 
In some cases, instead of giving the ``desired'' answer, the system generates an alternative, reasonable answer.
Our observations suggest that the system is able to deal with multi-turn generation better than an RNN counterpart and tends to be more consistent with respect to context  (Table~\ref{tab:generation_2}).\footnote{We provide a live invitation-only demonstration site for a conversational agents with toxicity controls and mutual information maximization features discussed in this paper. Check our GitHub repository for more information.}

\begin{table}[ht!]
\scriptsize
\small
\begin{tabular}{p{1.5in} |p{1.2in} }
\cmidrule[\heavyrulewidth]{1-2}
source & Response \\
\cmidrule[\heavyrulewidth]{1-2} 
who is the first president of the United States? & George Washington \\
what is the boiling point of water? & I think it's about 212 F. \\
which one is bigger, sun or moon? & The sun . \\
which animal has black and white stripes? & A black and white striped cat . \\
\end{tabular}
\caption{Addressing commonsense questions}\label{tab:generation}
\end{table}

\begin{table}[ht!]
\scriptsize
\small
\begin{tabular}{l |p{2.25in} }
\cmidrule[\heavyrulewidth]{1-2}
Role & response \\
\cmidrule[\heavyrulewidth]{1-2} 
User & Does money buy happiness? \\
Bot &  Depends how much money you spend on it .\\
User & What is the best way to buy happiness ? \\
Bot & You just have to be a millionaire by your early 20s, then you can be happy . \\
User & This is so difficult ! \\
Bot & You have no idea how hard it is to be a millionaire and happy . There is a reason the rich have a lot of money \\
\end{tabular}
\caption{An interactive example of multi-turn dialogue}\label{tab:generation_2}
\end{table}

\begin{table}[ht!]
\scriptsize
\small
\begin{tabular}{l |p{2.25in} }
\cmidrule[\heavyrulewidth]{1-2}
Role & Response \\
\cmidrule[\heavyrulewidth]{1-2} 
User & what is the meaning of life ? \\
Bot & The meaning is to be with your family and friends .\\
Bot & I'm going to guess : It means that your parents and friends have loved you . That's the meaning , I believe . \\
Bot & I think the meaning is to make sure you're alive when you die . So you'll be with your family , friends , and yourself .\\
Bot &It's not , but thanks for trying to help .\\
\end{tabular}
\caption{An example of multi-turn self-playing dialogue with user prompt}\label{tab:generation_3}
\end{table}

\subsection{Human Evaluation}

\begin{table*}[ht!]
\small
\centering

\begin{tabular}{r r | c| r l}
\cmidrule[\heavyrulewidth]{1-5}
 \multicolumn{5}{c}{\textbf{Relevance}: \textit{A and B, which is more relevant and appropriate to the immediately preceding turn?}}\\
\cmidrule[\heavyrulewidth]{1-5} 
\multicolumn{2}{c|}{System A} & Neutral & \multicolumn{2}{c}{System B}\\ 
\cmidrule[\heavyrulewidth]{1-5}
\cmidrule{1-5}
DialoGPT (345M) & 3281 (\bf{72}\%) & 394 (9\% ) & 882 (19\%) &   PersonalityChat ****\\
DialoGPT (345M) & 2379 (40\%) & 527 (9\% ) & 3094 (\textbf{52}\%) &  DialoGPT (345M, w/ MMI) ****\\
DialoGPT (345M) & 3019 (\bf{50}\%) & 581 (10\%)  & 2400 (40\%) & DialoGPT (345M, Beam) ****\\
DialoGPT (345M) & 2726 (\bf{45}\%) & 576 (10\%)  & 2698 (45\%) & DialoGPT (762M)\\
\cmidrule{1-5} 
DialoGPT (345M) & 2671 (45\%) & 513 (9\% ) & 2816 (\textbf{47}\%) &  Human response \\
DialoGPT (345M, w/ MMI) & 2871 (\bf{48}\%) & 522 (9\%) & 2607 (43\%) & Human response ***\\
\cmidrule[\heavyrulewidth]{1-5}
\end{tabular}

\begin{tabular}{r r | c| r l}
\cmidrule[\heavyrulewidth]{1-5}
 \multicolumn{5}{c}{\textbf{Informative}: \textit{A and B, which is more contentful, interesting and informative?}}\\
\cmidrule[\heavyrulewidth]{1-5} 
\multicolumn{2}{c|}{System A} & Neutral & \multicolumn{2}{c}{System B}\\ 
\cmidrule[\heavyrulewidth]{1-5}
\cmidrule{1-5}
DialoGPT (345M) & 3490 (\bf{77}\%) & 206 (5\%)  & 861 (19\% )& PersonalityChat ****\\
DialoGPT (345M) & 2474 (41\%) & 257 (4\%) & 3269( \textbf{54}\%) & DialoGPT (345M, w/ MMI) ****\\
DialoGPT (345M) & 3230 (\bf{54}\%) & 362 (6\%) & 2408( 40\%) & DialoGPT (345M, Beam) *****\\
DialoGPT (345M) & 2856 (\bf{48}\%) & 303 (5\%) & 2841( 47\%) & DialoGPT (762M)\\
\cmidrule{1-5} 
DialoGPT (345M) & 2722 (45\%) & 234 (4\%) & 3044( \textbf{51}\%) & Human response ****\\
DialoGPT (345M, w/ MMI) & 3011 (\bf{50}\%) & 234 (4\%) & 2755( 46\%) & Human response **\\
\cmidrule[\heavyrulewidth]{1-5}
\end{tabular}

\begin{tabular}{r r | c| r l}
\cmidrule[\heavyrulewidth]{1-5}
 \multicolumn{5}{c}{\textbf{Human-like}: \textit{A and B, which is more likely to be generated by human rather than a chatbot?} }\\
\cmidrule[\heavyrulewidth]{1-5} 
\multicolumn{2}{c|}{System A} & Neutral & \multicolumn{2}{c}{System B}\\ 
\cmidrule[\heavyrulewidth]{1-5}
\cmidrule{1-5}
DialoGPT (345M)& 3462 (\bf{76})\% & 196 (4\%) & 899  (20\%) & PersonalityChat ****\\
DialoGPT (345M)& 2478 (41)\% & 289 (5\%) & 3233 (\textbf{54}\%) & DialoGPT (345M, w/ MMI) ****\\
DialoGPT (345M)& 3233 (\bf{54})\% & 340 (6\%) & 2427 (40\%) & DialoGPT (345M, Beam) ****\\
DialoGPT (345M)& 2847 (\bf{47})\% & 321 (5\%) & 2832 (47\%) & DialoGPT (762M)\\
\cmidrule{1-5} 
DialoGPT (345M)& 2716 (45)\% & 263 (4\%) & 3021 (\textbf{50}\%) & Human response ***\\
DialoGPT (345M, w/ MMI)& 2978 (\bf{50})\% & 241 (4\%) & 2781 (46\%) & Human response *\\
\cmidrule[\heavyrulewidth]{1-5}
\end{tabular}

	\vspace{1mm}
\caption{Results of {\bf Human Evaluation} for relevance, informativeness and human-response possibility, showing preferences (\%) for our model (DialoGPT) vis-a-vis its variants and real human responses. Distributions skew towards DialoGPT with MMI, even when compared with human outputs. Numbers in bold indicate the preferred systems. Statistically significant results are indicated: * p $\leq$ 0.01, ** p $\leq$ 0.001, *** p $\leq$ 0.0001, **** p $\leq$ 0.00001. 
}\label{tab:human_eval}
\end{table*}

\paragraph{Human evaluations}

We evaluated 2000 randomly sampled test sources from the Reddit 6K test dataset using crowd-sourcing. Systems were paired and each pair of system outputs was randomly presented  to 3 judges, who ranked them for relevance, informativeness and how human-like the generating is using a 3-point Likert-like scale. Judges were required to pass a qualification test, and a regime of spam detection was imposed.\footnote{We used held-out hand-vetted data from the human and PersonalityChat datasets to provide clear-cut cases for spam prevention and judge training examples. We suspect that this may have helped bias the results towards the extremes.}
Overall judge preferences for relevance, informativeness and human-likeness, presented as raw numbers and a percentage of the total, are shown in Table \ref{tab:human_eval}. A strong preference can be observed for DialoGPT over PersonalityChat. Table \ref{tab:human_eval} also suggests that the "vanilla" DialoGPT medium model may already be close to human response quality. Unexpectedly, we found that judges may prefer the MMI variant over human responses, probably because of many of the true human responses are erratic or idiosyncratic, or are tied to internet memes that happened to be unfamiliar to the judges.\footnote{For example, one judge protested that the internet meme ``I was today years old when I realized this.'' did not seem human-like.} (See Section 4.2 for 
the conditions underlying this effect.)
Further details, including a test of significance and the human evaluation template used, are provided in the Appendix.

\section{Related work}
There are several open-sourced toolkits for large-scale pre-trained transformer models.
Huggingface Conv-AI transfer learning repository \cite{DBLP:journals/corr/abs-1901-08149} contains the code for training conversational AI systems with transfer learning based on the GPT-2 transformer language model, which achieves the state-of-the-art performance on ConvAI-2 dialogue competition. 
DLGnet \cite{olabiyi2019multi} is a large transformer model trained on dialogue dataset and achieves good performance in multi-turn dialogue generation.
AllenNLP \cite{allennlp} is developed as a toolkit for many natural language processing tasks, including the large-scale pre-trained bi-LSTM sentence representation learning framework ELMo \cite{elmo}.
Texar \cite{texar} focuses on text generation including style transferring and controllable generation. It includes reinforcement learning capabilities along with its sequence modelling tools.
DeepPavlov \cite{deeppavlov} is a popular framework focusing on task-oriented dialogue. This public repository contains several demos and pre-trained models for question answering and sentiment classification. 
Icecaps \cite{shiv2019microsoft} is a response generation toolkit with techniques such as grounding on personalities or external
knowledge and multi-task training.
The ConvAI2 challenge \cite{convai2} has a focus on personalized conversations.
ParlAI \cite{parlai} is another library for developing task-oriented dialogue systems. It contains pre-trained models for knowledge-grounded chatbot trained with crowd-sourced data.
The Text-to-Text Transformer \cite{2019t5} unifies multiple text modeling tasks, and achieves the state-of-the-art results in various natural language generation and understanding benchmarks.

\section{Limitations and risks}
\system is released as a model only; the onus of decoder implementation resides with the user. Despite our efforts  to  minimize  the  amount  of overtly  offensive  data  prior  to  training, \system retains the potential to generate output  that  may trigger  offense. Output  may reflect  gender  and  other  historical biases implicit in the data. Responses generated  using  this  model may exhibit a propensity to express agreement with propositions  that  are  unethical, biased or offensive (or the reverse, disagreeing with otherwise ethical statements). These are known issues  in  current state-of-the-art end-to-end  conversation  models trained  on  large  naturally-occurring datasets. A major motive for releasing \system is to enable researchers to investigate these issues and develop mitigation strategies. In no case should inappropriate content generated as a result of using \system  be construed to reflect the views or values of either the authors or Microsoft Corporation.

\section{Conclusion}

We have released an open-domain pre-trained model, \system, trained on massive real-world Reddit dataset. 
The package consists of a distributed training pipeline and several pre-trained models that can be fine-tuned to obtain a conversation model on a moderately-sized customized dataset in few hours. 
\system is fully open-sourced and easy to deploy, allowing users to extend the pre-trained conversational system to bootstrap training using various datasets. It serves as a building block to novel applications and methodologies. 
Detection and control of toxic output will be a major focus of future investigation. 
We will investigate leveraging reinforcement learning to further improve the relevance of the generated responses and prevent the model from generating egregious responses. 

\section*{Acknowledgements}

We would like to thank Yu Wang, Vighnesh Leonardo Shiv, Chris Quirk, and the anonymous reviewers for their helpful discussions and comments.

\bibliography{acl2019}
\bibliographystyle{acl_natbib}
\clearpage
\appendix
\onecolumn
\section{Additional Details of Human Evaluation }
\begin{figure}[h!]
    \centering
    \includegraphics[width=0.75\textwidth]{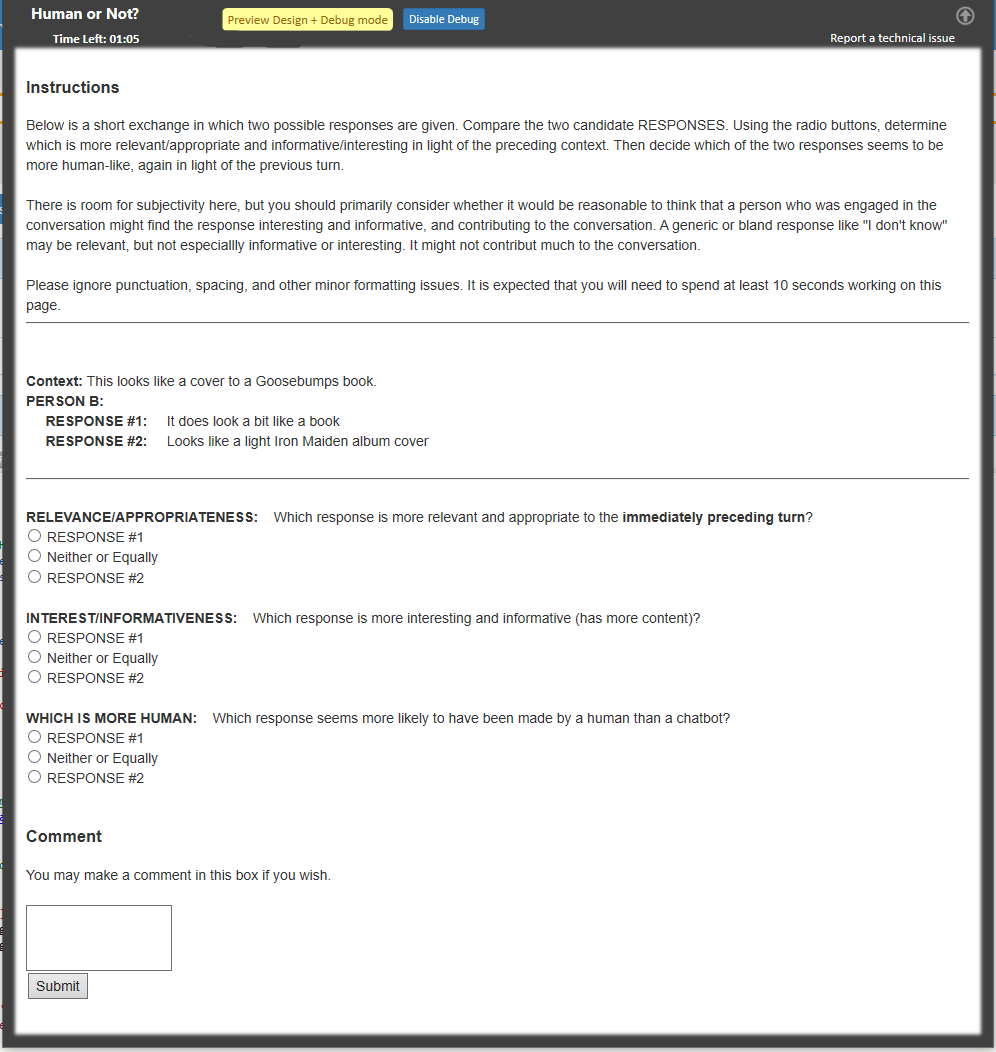}
    \caption{Human evaluation template}
    \label{fig:human_eval}
\end{figure}

Significance testing for the difference in means was performed using 10K bootstrap iterations. P-values are computed at $\alpha = 0.05$. The results are provided in Table~\ref{tab:sig}. The differences between 345M (2) and 762M (6) models are not significant. Notably also, the differences between 345M model (2) and human response (1) are not statistically significant. The template for human evaluation is provided in Figure~\ref{fig:human_eval}.
\begin{table*}[ht!]
\scriptsize
\begin{tabular}{c|ccc|ccc|ccc}
\hline
& System 1 &  &  & System 2 &  &  &   Pairwise &  \\
\hline
 & Mean & Std & 95\% CI & Mean & Std & 95\% CI & Std & 95\% CI & P-Value\\
2 vs 1 Human-like & 0.4527 & 0.0065 & ( 0.4400,  0.4653 ) & 0.5035 & 0.0065 & ( 0.4909,  0.5162 ) & 0.0127 & (-0.0758, -0.0259 ) & 0.0001 \\
2 vs 1 Informativeness & 0.4537 & 0.0065 & ( 0.4410,  0.4663 ) & 0.5073 & 0.0064 & ( 0.4948,  0.5199 )  & 0.0127 & (-0.0785, -0.0287 ) & 0.0000 \\
\textbf{2 vs 1 Relevance} & 0.4452 & 0.0064 & (0.4326,  0.4577 ) & 0.4693 & 0.0064 & ( 0.4568,  0.4819 ) & 0.0124 & (-0.0485,  0.0002 ) & 0.0552 \\
2 vs 3 Human-like & 0.7597 & 0.0064 & ( 0.7473,  0.7723 ) & 0.1973 & 0.0059 & ( 0.1858,  0.2089 ) & 0.0117 & ( 0.5392,  0.5852 ) & 0.0000 \\
2 vs 3 Informativeness & 0.7659 & 0.0063 & ( 0.7536,  0.7783 ) & 0.1889 & 0.0058 & ( 0.1777,  0.2003 ) & 0.0115 & ( 0.5540,  0.5993 ) & 0.0000 \\
2 vs 3 Relevance & 0.7200 & 0.1935 & ( 0.7070,  0.7333 ) & 0.1935 & 0.0067 & ( 0.7070,  0.7333 ) & 0.0117 & ( 0.5034,  0.5493 ) & 0.0000 \\
2 vs 4 Human-like & 0.4130 & 0.0063 & ( 0.4005,  0.4253 ) & 0.5388 & 0.0064 & ( 0.5263,  0.5514 ) & 0.0124 & (-0.1504, -0.1016 ) & 0.0000 \\
2 vs 4 Informativeness & 0.4123 & 0.0063 & ( 0.3999,  0.4246 ) & 0.5448 & 0.0064 & ( 0.5323,  0.5575 ) & 0.0124 & (-0.1570, -0.1082 ) & 0.0000 \\
2 vs 4 Relevance & 0.3965 & 0.0063 & ( 0.3841,  0.4088 ) & 0.5157 & 0.0064 & ( 0.5031,  0.5281 ) & 0.0122 & (-0.1431, -0.0955 ) & 0.0000 \\
2 vs 5 Human-like & 0.5388 & 0.0064 & ( 0.5263,  0.5513 ) & 0.4045 & 0.0063 & ( 0.3921,  0.4169 ) & 0.0125 & ( 0.1098,  0.1587 ) & 0.0000 \\
2 vs 5 Informativeness & 0.5383 & 0.0064 & ( 0.5258,  0.5508 ) & 0.4013 & 0.0063 & ( 0.3890,  0.4137 ) & 0.0124 & ( 0.1127,  0.1611 ) & 0.0000\\
2 vs 5 Relevance & 0.5032 & 0.0064 & ( 0.4906,  0.5157 ) & 0.4000 & 0.0063 & ( 0.3876,  0.4124 ) & 0.0122 & ( 0.079,  0.127 ) & 0.0000 \\
\textbf{2 vs 6 Human-like} & 0.4745 & 0.0065 & ( 0.4618,  0.4872 ) & 0.4720 & 0.0064 & ( 0.4596,  0.4846 ) & 0.0125 & (-0.0220,  0.0272 ) & 0.8476 \\
\textbf{2 vs 6 Informativeness} & 0.4760 & 0.0064 & ( 0.4634,  0.4887 ) & 0.4735 & 0.0064 & ( 0.4610,  0.4861 ) & 0.0126 & (-0.0221,  0.0273 ) & 0.8449 \\
\textbf{2 vs 6 Relevance} & 0.4543 & 0.0065 & ( 0.4417,  0.4671 ) & 0.4497 & 0.0064 & ( 0.4372,  0.4622 ) & 0.0123 & (-0.0193,  0.0289 ) & 0.7066 \\
4 vs 1 Human-like & 0.4963 & 0.0064 & ( 0.4838,  0.5090 ) & 0.4635 & 0.0065 & ( 0.4508,  0.4762 ) & 0.0127 & ( 0.0081,  0.0578 ) & 0.0094 \\
4 vs 1 Informativeness & 0.5018 & 0.0064 & ( 0.4894,  0.5144 ) & 0.4592 & 0.0127 & ( 0.0180,  0.0676 ) & 0.0127 & ( 0.0180,  0.0676 ) & 0.0009 \\
4 vs 1 Relevance & 0.4785 & 0.0064 & ( 0.4660,  0.4911 ) & 0.4345 & 0.0065 & ( 0.4218,  0.4472 ) & 0.0123 & ( 0.0199,  0.0682 ) & 0.0005 \\
\hline
\end{tabular}
\caption{Human evaluation significance test. Bold results represent differences that are \textbf{NOT} statistically significant. Notation: 1 - Human response; 2 - \system 345M; 3 - PersonalityChat; 4 - \system 345M w/ MMI; 5 - \system 345M Beam search; 6 - \system 762M }\label{tab:sig}
\end{table*}

\end{document}